# Braking and Body Angles Control of an Insect-Computer Hybrid Robot by Electrical Stimulation of Beetle Flight Muscle in Free Flight

T. Thang Vo-Doan, V. Than Dung, and Hirotaka Sato

*Abstract*—**While engineers put lots of effort, resources, and time in building insect scale micro aerial vehicles (MAVs) that fly like insects, insects themselves are the real masters of flight. What if we would use living insect as platform for MAV instead? Here, we reported a flight control via electrical stimulation of a flight muscle of an insect-computer hybrid robot, which is the interface of a mountable wireless backpack controller and a living beetle. The beetle uses indirect flight muscles to drive wing flapping and three major direct flight muscles (basalar, subalar and third axilliary (3Ax) muscles) to control the kinematics of the wings for flight maneuver. While turning control was already achieved by stimulating basalar and 3Ax muscles, electrical stimulation of subalar muscles resulted in braking and elevation control in flight. We also demonstrated around 20 degrees of contralateral yaw and roll by stimulating individual subalar muscle. Stimulating both subalar muscles lead to an increase of 20 degrees in pitch and decelerate the flight by 1.5 m/s² as well as an induce an elevation of 2 m/s².**

*Index Terms*— **Insect-Computer Hybrid Robot, MAV, Biological-inspired Robot, Insect Flight, Flight Control, Insect Flight Muscle**

## I. Introduction

Engineers have been developing robots by mimicking animal locomotion for decades, but it is still challenging to match the performance capability of the animals, especially insects, in terms of actuating, sensing, energy efficiency, and locomotion control [1]–[9]. Scaling the robots to insect size also faces a bottle neck in durability of materials (e.g., actuators, chassis) that limit their life cycle significantly [3], [10]. Meanwhile insects are not only able to walk, run, and climb smoothly on complex terrains but also perfect flyers. Using live insects as platforms for developing robots would inherit all the insect capability and properties while bypassing complex mechanical and material design as well as fabrication process. Insect-computer hybrid robot is the interface of a living insect platform and a mounted control backpack. The motor actions of the insect can be controlled by sending the electrical stimulus directly from the backpack terminals to the muscles or neural clusters through implanted electrodes. Precise walking gaits in beetles was achieved by stimulating the leg muscles [11], [12] while turning, backward, forward, and sideways walking were driven by stimulating the mechanosensory organs (e.g., antennae, cercus, and elytra) and ganglion in beetles [13], [14] and cockroaches [15]–[17]. Flight initiation and cessation of the beetles were achieved by stimulating optic lobes [18] and indirect flight muscles [19] while stimulating direct flight muscles enable steering control in flight [18], [20], [21]. In addition, there were also attempts to control moth flight by direct stimulation of insect brain areas [22], ventral nerve cord that controls abdominal deflection [23] and antennal muscles [24].

Insect-scale micro aerial vehicles (MAVs) has drawn lots of interest in structure and material design and optimization toward untethered autonomous flight as well as endurance lifetime. While it took engineers years or even decades to achieve a perfect insect-scale MAV [1], [3], [10], [25]–[29], insects are already perfect flyers. They consist of indirect flight muscles for flapping the wings, direct flight muscles for altering wing kinematics, and resilient cuticle of thorax and wing hinges that can last for a lifetime. They also have a complex sensory and neural network for precise feedback control of motor actions. Insects thus have high maneuverability in flight as they can easily turn, adjust altitude, body angles, hover, and even have extreme motion like saccade [30]–[36]. Insect-computer hybrid robot based on living beetle platform is a preferred model for alternative MAV due to not only its high maneuverability but also its high payload in flight and endurance cuticle to protect them from crashing [37]. While stimulating basalar muscles (Fig.1) increase the wing beat amplitude that cause contralateral turn [18], the activation of third axillary muscle (3Ax) resulted in ipsilateral turn due to decrease in wing amplitude [20]. In addition, stimulating basalar and 3Ax muscle alternatively with a feedback control

This work was supported by the Singapore Ministry of Education [grant number: MOE2017-T2-2-067]. (Corresponding author: Hirotaka Sato.)

T.T. Vo-Doan was with School of Mechanical and Aerospace Engineering, Nanyang Technological University, Singapore. He is now with the Institute of Biology I, University of Freiburg (e-mail: vodoan@bio.uni-freiburg.de).

V.T. Dung was with School of Mechanical and Aerospace Engineering, Nanyang Technological University, Singapore. He is now with Fossil Vietnam (e-mail: dvthan@gmail.com).

H. Sato is with School of Mechanical and Aerospace Engineering, Nanyang Technological University, Singapore (e-mail: hirosato@ntu.edu.sg).

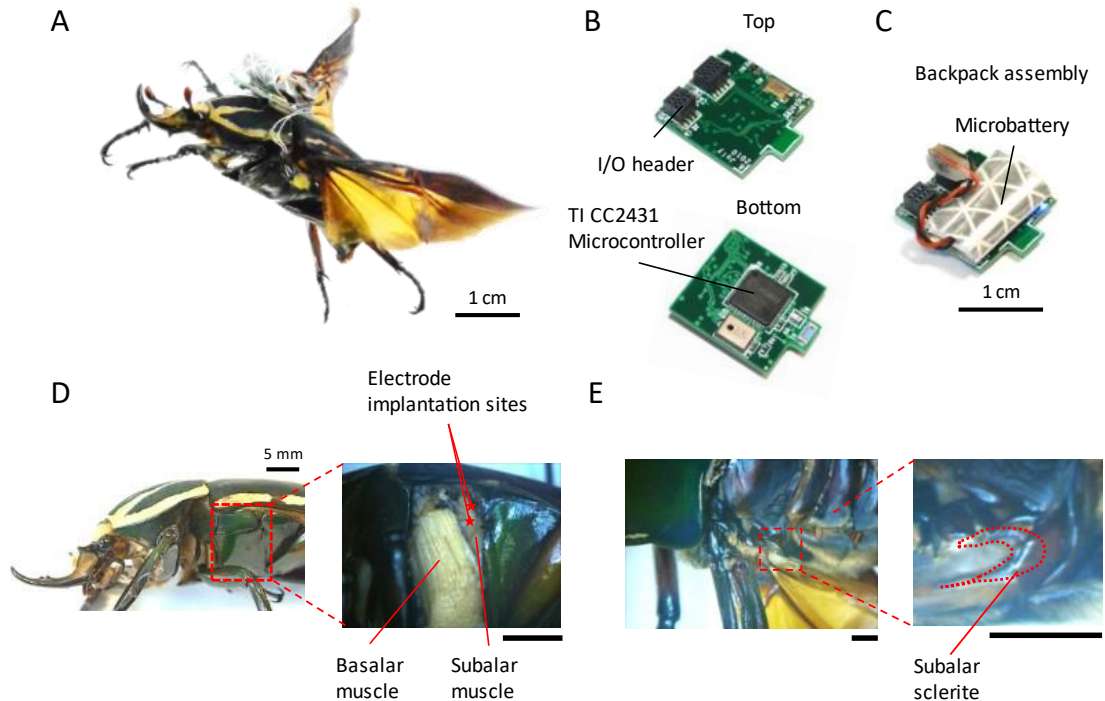

Fig. 1. Overview of an insect-computer hybrid robot. (A) Live beetle with mounted backpack. (B) The wireless backpack (electrical stimulator) was customized using the TI CC2431 microcontroller with the tiny I/O header for connecting the electrodes and a microbattery. (C) The battery was wrapped with retro-reflective tape and mounted on the top of the backpack to supply power and serve as a marker for motion tracking. (D) Lateral view of the beetle with a close-up view of the direct flight muscles after dissecting the cuticle. The subalar muscle is located in the posterior part of the thorax, next to the basalar muscle. It runs from the hind leg coxa to an apodema that connects to the subalar sclerite. (E) Dorsal view of the beetle, after removal of the elytra. Exposed view of the subalar sclerite showing connection to the wing base via a flexible membrane.

system could drive the beetle to follow a predetermined path in flight [21]. Although turning control of beetle flight is well established, the control of body angles and braking is lacking. Such capability of body angle and braking control would enable more complex maneuverability of beetle flight such as hovering and saccade.

Here, we demonstrated a stimulation protocol of subalar muscle, the last major direct flight muscle besides basalar and 3Ax muscles, for control the braking and body angles of an insect-computer hybrid robot based on a live beetle (Coleopteran, Mecynorrhina torquata) in flight (Fig. 1A-C). During fictive decelerated flight in tethered condition, the firing rate of subalar muscle and the wing rotation angle increased which suggested the role of subalar muscle in braking and pitch control. Stimulating subalar muscle in free flight induce contralateral yaw and roll of the body and increase in pitch angles. Moreover, stimulating both subalar muscles simultaneously also resulted in braking and elevation in flight.

## II. MATERIALS AND METHODS

### A. Animals

Adult Mecynorrhina torquate beetles (6 cm – 8 cm, 8 – 10 g) were imported from Kingdom of Beetle, Taiwan and reared in separate containers (20 × 8 × 12 cm) with woodpile beading. They were fed sugar jelly every 3 days. The temperature and humidity of the rearing room was maintained at 25°C and 60%. Invertebrates, including beetles, does not require ethics approval for animal research according to the National Advisory Committee for Laboratory Animal Research.

### B. Electrode Implantation

Beetles were anesthetized for 3 min in a small $CO_2$-filled container. The beetle was then placed on a wooden block and gently covered with dental wax, which was pre-softened by dipping into hot water. Two silver wire electrodes (diameter of the bare wire, 127 µm and diameter of the coated wire, 178 µm; A-M Systems) were implanted into the subalar muscle at a depth of 3 mm through a small opening on the cuticle. Before the electrodes were inserted, the bare silver was exposed by flaming. The electrodes were secured in place by melted beeswax.

### C. Wireless Backpack Assembly

The wireless system included two electronic devices that were based on the TI CC2431 microcontroller [20]. One was the miniature wireless backpack, which was mounted onto the beetle and the other was the base station connected to the computer. The backpack was driven by a micro poly lithium ion battery (Fullriver, 3.7 V, 350 mg, 10 mAh) wrapped with retro-reflective tape (Silver-White, Reflexite), which also served as a marker for the motion capture system. The total mass of the backpack, including the battery, was $1.2 \pm 0.26$ g. The additional mass from the backpack has minimal effect on beetle flight speed. The backpack mounted beetles behaved as close as load-free beetle while excessive load significantly decreased the velocity of the beetle flight (Fig. 1 and S1).

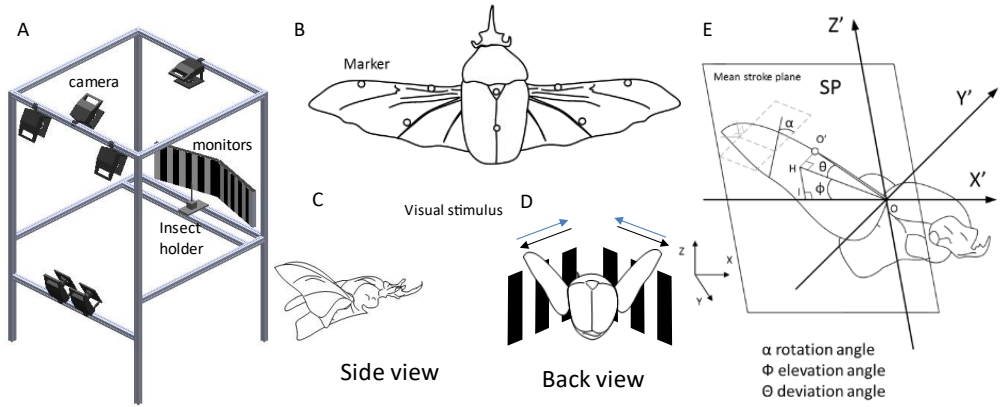

Fig. 2. Tethered experiment setup. (A) Six near-infrared cameras were mounted onto an aluminum frame to cover a volume of $400 \times 400 \times 400$ mm surrounding the beetle holder at a frame rate of 1500 fps. The beetle was fixed on the holder with the head towards the two monitors (placed at 250 mm from the beetle head) to see the moving pattern generated. For visual stimulation, the patterns were 20 mm black/white horizontal stripes moving forward and backward at 50 Hz within the screen. During electrical stimulation, the moving stripes were changed to a still white background on the screen. (B) Each wing was attached to three retro-reflective tape markers ($2 \times 2$ mm) for reconstructing the wing kinematics and two markers on the body for reference. (C-D) Lateral and posterior view of the beetle presented visual stimulus of vertical stripes from the panel. (E) The wing position was defined with a 60° tilted body axis after rotating the wing stoke plane (SP) to the horizontal plane for calculation. The three Euler angles of the wing ($\phi$, elevation; $\theta$, deviation; and $\alpha$, wing rotation angles) were defined. The elevation angle is the wing stroke angle measured about the Y' axis in the mean stroke plane. The deviation angle is the out-of-plane angle that the leading edge of the wing makes with the stroke plane, the wing rotation angle is the angle the wing cord makes with the plane created by the leading edge and the Y' axis, and (X' Y' Z') is the wing stroke plane frame.

An IMU backpack was used to record the body angles and accelerations of the flying insect. The backpack contains a nine axis IMU MPU9250 from Invensense to measure orientation and movement of the board and the BLE microcontroller CC2642R from Texas Instrument. The IMU integrates a small digital computational engine to do sensor fusion from its nine axes of data sensor which 3-axis acceleration, 3-axis gyroscope, and 3-axis magnetometer. The IMU sensor results in three linear accelerations and the sensor orientations in quaternion format. The IMU was sampled with sampling rate 100Hz. The total mass of the backpack, including the battery, was $1.6 \pm 0.26$ g.

### D. Tethered Experiment

The beetle was placed within the field of view of six near-infrared cameras (VICON T40s) in the three-dimensional (3D) motion capture system (Fig. 2A) to record wing kinematics. Retro-reflective markers are attached to the wings and body of the beetle (Fig. 2B), then the silver wire electrodes were implanted into the subalar muscle and connected to the EMG recorder (CC2431 Microcontroller Development board) through an instrument amplifier (Analog Devices, LT1920) or a function generator. For visual stimulation, a visual stimulator was set to project the optic flow pattern (vertical black/white stripes of 20 mm) onto two monitors placed 250 mm in front of the beetle (Fig. 2C-D). When the patterns moved, the beetle changed the wing kinematics to track the moving patterns. The

EMG signal was recorded from the subalar muscle and synchronized with the wing kinematics (Fig. 2E). For electrical stimulation, the wing kinematics were synchronized with the electrical stimulus.

### E. Free Flight Experiment

After assembly, the wireless connection to the backpack and operation laptop was set up. The beetle was then released into the flight arena and stimulated via a Wii remote. Activation of the Wii remote sent a command to the BeetleCommander running on the laptop, which sent a wireless command to the backpack via the base station. After receiving the command, the backpack generated the electrical stimulus signal and applied it to the subalar muscle. The beetle position was recorded simultaneously by the 3D motion capture system or the IMU data is then synchronized with the electrical stimulation.

A 500 ms pulse train was applied on the subalar muscle in free flight for electrical stimulation. The pulse amplitude and pulse width are 3V and 3 ms, respectively (Fig. S2). The stimulation frequency is varied from 40 Hz to 100 Hz. The stimulation parameters were selected from a brief survey at the beginning of the experiment setup and limited to the most effective range due to the constrain of flight duration of the beetles. Besides, the electrical stimulation of subalar muscle does not affect other nearby muscles (Fig. S3).

The free flight trajectories of the beetles were sorted to individual flight paths of 150 ms before stimulation and 500 ms during stimulation. The x, y, z components of the velocity and acceleration were calculated after smoothing the flight path using a 5th order polynomial function by a customized MATLAB code ($N_{both} = 14$ beetles, $n_{both} = 495$ trials, $n_{left} = 360$ trials, $n_{right} = 346$ trials). Saccade flight paths whose angular velocities were over 500°/s were excluded from the data analysis because such a fast motion could not be accurately detected. The horizontal and lateral vectors were calculated for every pair of points along the flight path on the (x, y) plane. The horizontal and lateral accelerations were calculated by summing the projected x and y accelerations relative to the horizontal and normal vectors, respectively. The vertical acceleration is the z component of the acceleration.

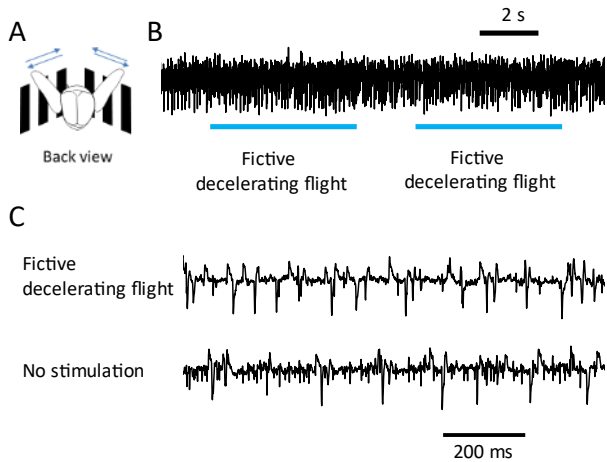

Fig. 3. Electromyography of the beetle during visual stimulation. (A) The beetle was mounted in front of two monitors (form an angle of 120 degrees) for visual stimulation. Vertical black/ white stripes moved at around 10 Hz to emulate forward/ backward visual stream. A pair of silver electrodes were implanted into the subalar muscle for recording EMG signal. (B) The subalar muscle fired continuously during beetle flight. (C) It was most active during fictive decelerating flight, with a firing rate of 25.3 Hz, which decreased to 21.6 (p < 0.05, paired samples t-test) during the period of no stimulation.

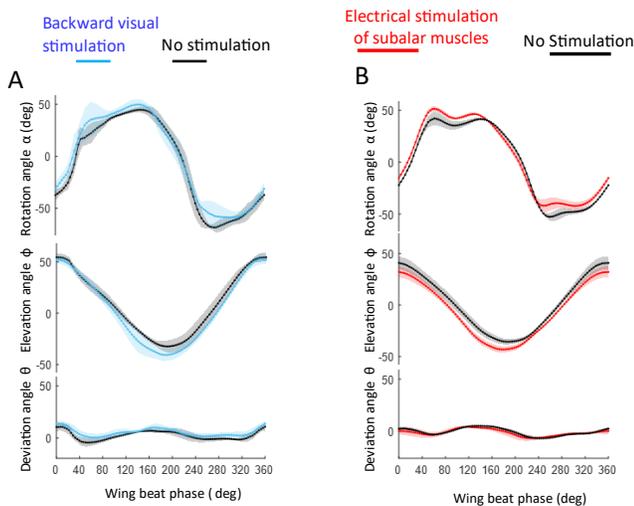

Fig. 4. Wing beat trajectories recorded during visually stimulated fictive decelerating flights (N = 4 n = 20 trials) and electrical stimulation of subalar muscle (N = 4, n = 98 trials). (A) The backward visual stimulation induced a positive phase shift with a 10° increase (p < 0.05, paired samples t-test) in the wing rotation angle from 80° to 180° and from 220° to 300° of the wing cycle. No change was observed in the deviation angle (p > 0.05, paired samples t-test) while the wing elevation angle showed a reduction of 10° from 80° to 180° of the wing cycle. The insert represents the magnification of the wing rotation angle at the phase with the highest change. (B) The beetle showed a slight phase shift during the first 80° of the wing cycle and a clear increase of 10° (p < 0.05, paired samples t-test) in wing rotation from 80° to 180° and from 220° to 300° of the wing cycle when the subalar muscle was stimulated. No change was observed in wing deviation angle while there is a negative shift of 10° in wing elevation angle (N = 4, n =98 trials).

The IMU data was collected from different beetles (N = 10 beetles, $n_{both}$ = 573 trials, $n_{left}$ = 277 trials, $n_{right}$ = 207 trials). A 5th order Butterworth filter with cutoff frequency of 20 Hz was applied for acceleration data. The induced amount is defined as the difference of the peak value within first 300 ms during stimulation and the onset value.

### F. Statistical Analysis

The pared samples t-test was performed to check the significance of the beetle responses before and during stimulation. The one simple t-test was used to check if the induced acceleration or angle was significant. Spearman correlation test was used to check the significance of the graded responses. A p value less than 0.05 was considered statistically significant.

## III. RESULTS AND DISCUSSIONS

### A. The Role of Subalar Muscle

The beetle subalar muscle is inserted from the hind leg coxa to the apodema of the subalar sclerite [32], [36]. The subalar muscle lies next to the basalar muscle at the posterior end of the thorax (Fig. 1 D-E). Its function would be similar to those of the III1 muscle of the blowfly (41, 42), M99 and M129 muscles of the locusts (26, 43–45) due to similarity in configuration. Anatomically, the contraction of the subalar muscle pulls the posterior part of the wing base, consequently, depresses the trailing edge of the wing that increases the wing rotation angle and thus the wing angle of attack. This would lead to a decrease in thrust and increase in lift generated by the wings of the flying insect [38]–[41]. Such change in thrust and lift of the wings would alter the body angles or induce brake in flight [33], [38]–[41][39], [40], [42].

The beetle subalar muscle was innervated in a graded fashion under tethered condition. It was more active during fictive decelerating flight than non-stimulated flight (Tethered Experiment section). During flight, we observed that the subalar muscle was continuously active (Fig. 3A). We provided optic flow stimuli to induce fictive decelerating flight in tethered beetles [35], [41], [43]. The firing rate of the electromyogram (EMG) spike during fictive decelerating flight was higher than that during no stimulation (Fig. 3B) (p < 0.05, paired samples t-test). This would suggest that the beetle increase the activity of subalar muscle during braking in flight.

Whether stimulated due to fictive backward visual stimuli (Fig. 4A), or via direct electrical stimulation (Fig. 4B), enhanced activity in the subalar muscle caused increase in wing rotation angle. Compared to spontaneous flight (black traces, Fig. 4A), the wing rotation and elevation angles respectively increased during fictive decelerating flight, with only slight changes in the corresponding deviation angle (N = 4, n = 20, p < 0.05, paired samples t-test). Similar to braking in flight, electrical stimulation of subalar muscle led to the increment in wing rotation and elevation angles (Fig. 4B) while little change was observed in the deviation angles (N = 4, n = 98, p < 0.05, paired sample t-test). These results would suggest that the beetle activates subalar muscle to increase wing rotation angle when braking in flight.

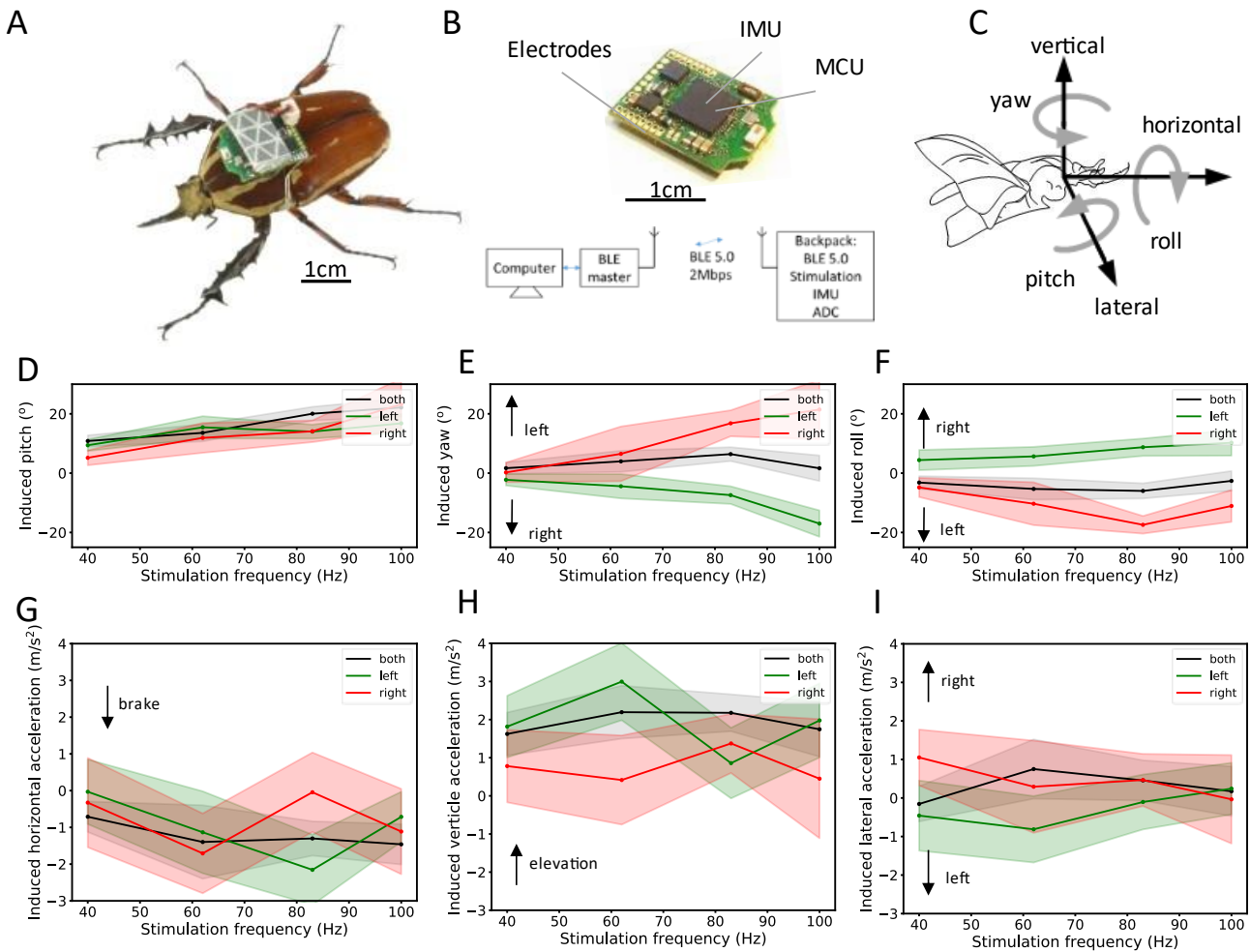

Fig. 5. Remote electrical stimulation with IMU backpack during free flight. (A) The IMU backpack mounted on the beetle. (B) The backpack consists of a Bluetooth low energy (BLE) CC2642R microcontroller and a nine-axis inertial measurement unit (IMU) MPU9250 from Invensense. The backpack processes the commands from control program to execute control signals, collect and return data. A computer communicates with the backpack via BLE 5.0 interface to transfer all control signal and collect and store all backpack information including stimulation signal and IMU data. The experiment was conducted in a flight arena of $12 \times 8 \times 4$ m. (C) The positive direction of the accelerations and body angles. (D) The horizontal acceleration of the beetle was decreased in all cases and was graded by increasing the stimulation frequency with a weak correlation coefficient of - 0.1 (p=0.014, Spearman correlation test) when both muscles were stimulated. There is no correlation of the horizontal acceleration and the stimulation frequency when individual muscle was stimulated. (E) The induced vertical acceleration remained positive when stimulating both subalar muscles (p<0.05, one sample t-test). (F) There is slight ipsilateral increase in lateral acceleration when individual subalar muscle was stimulated (p<0.05, one sample t-test). (G) The induced pitch angle of the beetle increased in all cases with a weak correlation coefficient of 0.23 when graded by stimulation frequency (p=0.013, Spearman correlation test). (H) The induced yaw angle increased contralaterally when stimulating individual subalar muscle with a weak correlation coefficient of 0.26 (p=0.00017, Spearman correlation test). (I) The induced roll angle increased contralaterally when stimulating individual subalar muscle with a weak correlation coefficient of 0.19 (p=0.0049, Spearman correlation test). Black, green, and red lines represent the results of electrical stimulation of both, left and right subalar muscle, respectively. Shaded regions indicate 95% confidence interval.

### B. Body Angles Control of the Insect-Computer Hybrid Robot in Free Flight

In free flight, the beetle with mounted IMU backpack enabled us to record body angles and accelerations of the insect during flight (Fig. 5A-C). Stimulating either left or right subalar muscle also result in increase in pitch angle from 5 degrees to 22 degrees with a weak correlation with stimulation frequency (coefficient = 0.25, p < 0.0001, Spearman correlation test) (Fig. 5D). The increase in pitch angle was from 10 degrees to 22 degrees (coefficient = 0.23, p = 0.013, Spearman correlation test) when both subalar muscles were stimulated while there is slight change in yaw and roll angles (Movie S1). Such increase in pitch angle would be the result of the induced lift of the wing during the stimulation which is similar to that found in Drosophila [39], [41], [44]. Stimulating individual subalar muscle lead to graded contralateral increase in yaw angle from 2 degrees to 17 degrees (coefficient = 0.26, p = 0.00017, Spearman correlation test) and roll angle from 5 degrees to 10 degrees (coefficient = 0.19, p=0.0049, Spearman correlation test) when the stimulation frequency was tuned from 63 Hz to 100 Hz (Fig. 5E-F).The change in body yaw and roll angles would be due to the asymmetry of the wing rotation angles during the stimulation. The electrical stimulation of subalar muscle would increase the wing rotation angle that lead to the increase in lift of that side [40]. Such increases are associated with the increase in vertical acceleration and decrease in horizontal acceleration of the beetle during flight.

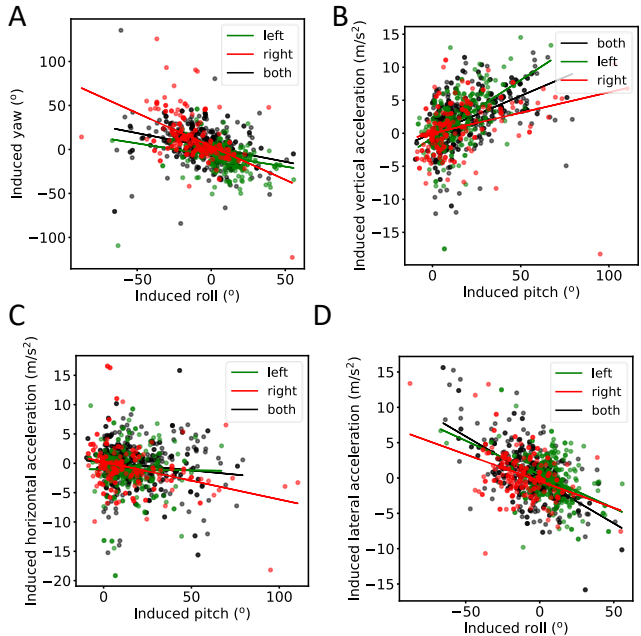

Fig. 6. The correlation of induced accelerations and body angles. (A) The induced yaw and roll angle have a moderate correlation coefficient of -0.41 (p < 0.0001, Spearman correlation test). (B) The induced horizontal acceleration has a weak correlation coefficient of -0.13 with the induced pitch angle (p=0.0013, Spearman correlation test). (C) The induced lateral acceleration has a weak correlation coefficient of -0.29 with the induced roll angle (p < 0.0001, Spearman correlation test). (D) The induced vertical acceleration has a moderate correlation coefficient of 0.49 with the induced pitch angle (p < 0.0001, Spearman correlation test).

### C. Braking Control of the Insect-Computer Hybrid Robot in Free Flight

A decrease of 0.7 m/s² to 1.4 m/s² of horizontal acceleration (Fig. 5G) and increase of 1 m/s² to 1.6 m/s² of vertical acceleration (Fig. 5H) were observed when stimulating both subalar muscles (Movie S1). The increase in induced lateral acceleration of 0.5 m/s² to 1 m/s² when individual subalar muscle was stimulated would be a side slip (Fig. 5I). There is decrease in horizontal acceleration and increase in vertical acceleration when individual subalar muscle was stimulated but without a clear trend of change (Fig. 5 G-H). In addition, the induced yaw angle has a moderate negative correlation with the induced roll angle (coefficient = 0.49, p < 0.0001, Spearman correlation test), which is similar to that of a bank turn observed in Drosophila [33], [45] (Fig. 4 A). The increase in pitch angle is associated with the elevation during flight (correlation = 0.49, p < 0.0001, Spearman correlation test) (Fig. 6B). The induced horizontal acceleration and lateral acceleration have a weak correlation coefficient of -0.13 and -0.29 with the induced pitch and induced roll, respectively (p = 0.0013, Spearman correlation test) (Fig. 6C-D). These correlations are also in agreement with those observed in Drosophila [39], [41]. These findings support our hypothesis that the subalar muscles have a function of controlling brake, elevation, and body angles of the beetles during free flight.

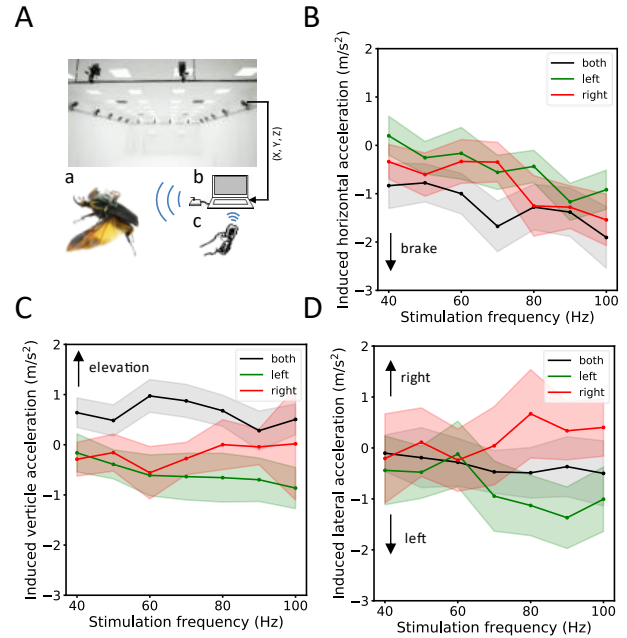

Fig. 7. Remote electrical stimulation during free flight. (A) The experiment was conducted in a flight arena of 12 × 8 × 4 m covered by a motion capture system equipped with 20 near-infrared cameras (VICON, T40s, and T160). (a) The backpack was assembled and mounted onto the beetle before releasing it into the air for free flight. The backpack wirelessly received the command from the operator's laptop via the base station (b) plugged into the laptop when the operator pressed the command button of the Wii remote (c). The backpack then applied the electrical stimulus to the subalar muscles. Meanwhile, the positions of the flying beetle were recorded with timestamps by the motion capture system and fed into the laptop for synchronizing with the stimulation command. (B) The horizontal acceleration of the beetle was decreased in all cases and reached up to approximately 2 m/s2 when stimulating both subalar muscles at 100 Hz. The horizontal acceleration was graded by increasing the stimulation frequency with a weak correlation coefficient of - 0.25 (p < 0.0001, Spearman correlation). (C) The induced vertical acceleration remained positive when stimulating both subalar muscles (p < 0.05, one sample t-test). Stimulating individual subalar muscle caused reduction in vertical acceleration (p < 0.05, one sample t-test). (D) The induced lateral acceleration shifted toward left with an amount of up to 0.4 m/s2 (p < 0.05, one sample t-test) when stimulating both subalar muscles. Stimulating individual muscle lead to increase up to 1.5 m/s2 ipsilaterally for the stimulation frequency from 70 Hz (p < 0.05, one sample t-test). Shaded regions indicate 95% confidence interval.

In addition, recorded positions of freely flying beetles showed the same tendency observed with IMU backpack. The stimulation of the subalar muscles leaded to the decrease in horizontal acceleration and increase in vertical acceleration (Fig. 7A-B). The induced acceleration was computed from the free flight path data (refer to "Free Flight data analysis" and [20]). During free flight, we also found that the induced horizontal acceleration was negative (Fig. 7B) and gradually reduced to −1.6 m/s² with a moderate dependence on the stimulation frequency (p < 0.0001, Spearman correlation). In addition, stimulation of the subalar muscle induced a relatively small but significant 0.5 m/s² to 1 m/s² increase in the induced vertical (p < 0.05, one sample t-test) (Fig. 7C). Stimulate left or right subalar muscle caused an increase in induced lateral acceleration of about 0.8 m/s² to 1.5 m/s² ipsilaterally when the stimulation frequency is within 70 Hz to 100 Hz ($p < 0.05$, one sample t-test) (Fig. 7D).The decrease in the horizontal directions and increase in the vertical directions of the induced accelerations indicate the increase in drag and lift of the wings.

Our data showed that the role of subalar muscle in beetle flight is different from that of basalar and 3Ax muscles. While the activation basalar and 3Ax muscles cause increment and reduction in wing beat amplitude, respectively[18], [20], activating subalar muscle lead to an increase in wing rotation angle. In addition, we found that activating subalar muscle in free flight induced change in body angles and braking which are distinct from the steering role of basalar and 3Ax muscles.

### D. Toward Autonomous Flight of Insect-machine Hybrid Robot

Although there are attempts to implement feedback control for freely flying beetles, it is still a challenge to realize autonomous flight of insect-machine hybrid robot [21], [46]. It would be useful to have a detail physiology investigation on the role of individual flight muscles in control of wing kinematics [47]–[49]. Such data would provide more insight of how the individual flight muscles work and their effect in flight and thus suggest appropriate stimulation protocol to control the flight muscle. For instance, the phasic activation pattern of a muscle would suggest the efficient stimulation timing of such muscle.

Recording activities of multiple muscles and high-resolution images of the wings in tethered condition is convenient and would provide certain relationship of muscle activation and wing kinematics [47], [48], [50], [51]. However, the behavior of insects in such constrained condition would differ from freely moving ones and it is impossible to achieve complex behaviors of the insect such as hovering and perching in such condition. While a large-scale virtual reality system can help to reproduce complex behaviors of freely flying insect [52], free flight EMG recording of flight muscles [53]–[55] along with high-resolution recording of the wings and body [33], [56]–[58] would provide precise correlation of muscle activation patterns and flight maneuvers. Implementing such data into a feedback control system would enable autonomous flight of insect-machine hybrid robot in the future.

### IV. CONCLUSION

Fully control the beetle flight requires the ability to activate multiple flight muscles. In this paper, we demonstrated that the role of subalar muscle in manipulating wing rotation angle which differs from regulating wing beat amplitude of basalar and 3Ax muscles. We were then able to control braking and lift as well as body angles of a beetle in free flight by activating subalar muscle. A feedback control system with the capability to stimulate all direct flight muscles would enable to achieve complex maneuver control of the beetle flight.

### ACKNOWLEDGMENT

The authors would like to thank Mr. Duc Long Le, Mr. Huu Duoc Nguyen, and Dr. Li Yao for the discussion, Mr. Roger Tan Kay Chia, Mr. Chew Hock See, and Mr. Seet Thian Beng for their support.

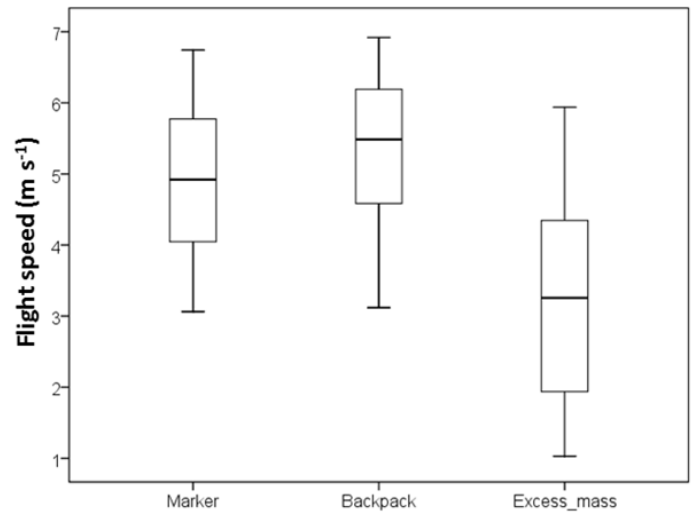

**Fig. S1. The flight speed of beetle with different loads.** The beetles experienced 3 different loading conditions (only marker mounted (0.25 g), backpack mounted (1.23 g) and excess mass mounted (3.50 g)) while flying freely in the flight arena.  There is no difference in flight speed of the beetles mounted only small marker and those with backpack (p>0.05, paired samples t-test) while those mounted excess mass show clear reduction in flight speed (p <0.001, paired samples t-test) (N=4 beetles, n=60 trials).

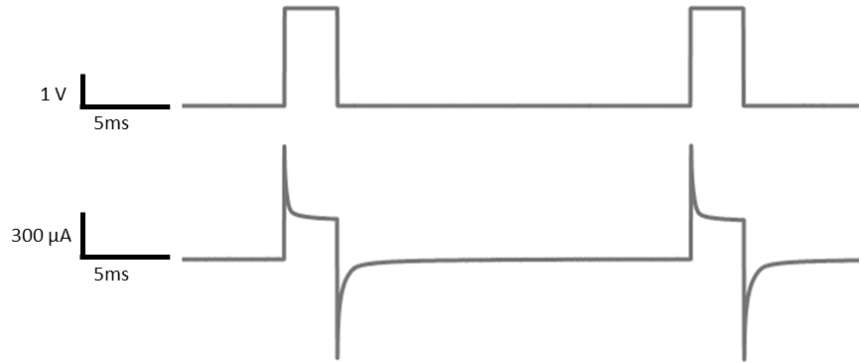

**Fig. S2.** Stimulation waveform and electrical current flow through the subalar muscle. (a) The stimulation pulse train (3V, 3 ms pulse width and 50 Hz) from the waveform function generator was applied on the subalar muscle via the two implanted electrodes. (b) The current waveform passed through the subalar muscle.

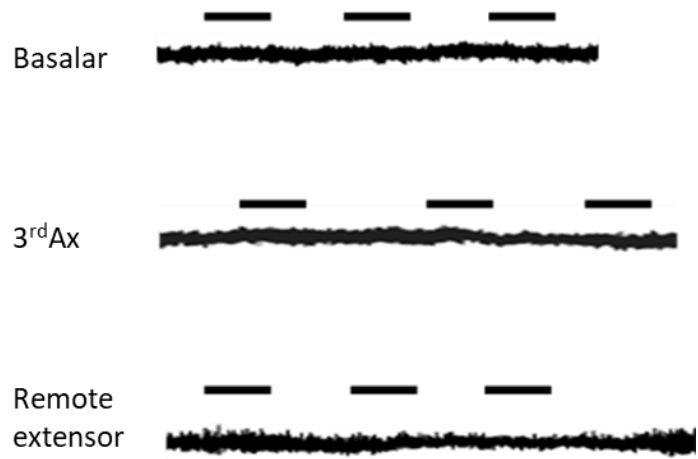

**Fig. S3. EMG of nearby muscles during the electrical stimulation of the subalar muscle.** One pair of silver wires was implanted into the subalar muscle for electrical stimulation another pair of silver wires was implanted to basalar muscle, $3^{rd}$Ax muscle or remote extensor muscle for recording EMG signal. Except the nerves, all the hinges and joints were removed before recording to avoid mechanical interaction of the muscles. No EMG spikes observed in the nearby muscles while stimulating the subalar muscles (N = 3, n = 60, p < 0.05, binomial test).

**Movie S1.** The response of the beetle due to electrical stimulation of subalar muscle. When both subalar muscles were stimulated, the beetle pitched and climbed up. It then recovered the body angle after the stimulation finished. The red and blue bars represent the body axis of the beetle with and without electrical stimulation, respectively.